\documentclass[10pt,twocolumn,letterpaper]{article}

\usepackage{cvpr}              

\usepackage{soul}

\definecolor{cvprblue}{rgb}{0.21,0.49,0.74}
\usepackage[pagebackref,breaklinks,colorlinks,allcolors=cvprblue]{hyperref}

\def\paperID{*****} 
\def\confName{CVPR}
\def\confYear{2026}

\title{On the Limits of Token Reduction for Efficient Unified Vision Language Training}

\author{
Siyi Chen$^{1}$ \thanks{Work done during an internship at Sony AI.} \quad 
Weiming Zhuang$^{2}$ \quad 
Jingtao Li$^{2}$ \quad 
Lingjuan Lv$^{2}$ \\
\\[-2pt]
$^{1}$University of Michigan \qquad
$^{2}$Sony AI 
}

\begin{document}
\maketitle
\begin{abstract}

Unified vision-language models (VLMs) integrate visual understanding and visual generation within a single autoregressive backbone, but their joint training is computationally expensive and largely overlooked from an efficiency perspective. In this work, we study the feasibility and limits of token-reduction-based acceleration for unified VLM training. Through a systematic analysis of layerwise attention allocation, we uncover a fundamental asymmetry: visual understanding exhibits substantial late-layer visual redundancy, whereas visual generation maintains persistent dependence on image tokens across depth. Guided by this observation, we design task-specific accelerators that selectively reduce image-token computation for each objective. While these methods achieve significant efficiency gains in isolated settings, we observe a consistent synergy loss under unified training—task-specific token dropping necessitates divergent parameter pathways and eliminates the mutual performance gains typically observed in joint optimization. Our findings suggest that efficient unified modeling requires preserving shared cross-task structures, highlighting the need for synergy-aware acceleration strategies.
Project page: \url{https://chicychen.github.io/TokenReductionUnifiedVLM/}.
\end{abstract}
    
\vspace{-.15in}
\section{Introduction}
\label{sec:intro}

Unified Vision-Language Models (VLMs) \cite{ma2024janusflowharmonizingautoregressionrectified,liquid,wang2024emu3nexttokenpredictionneed,wu2024vila,ma2025unitokunifiedtokenizervisual,ho2020denoisingdiffusionprobabilisticmodels} integrate visual generation \cite{tian2024visualautoregressivemodelingscalable,esser2024scalingrectifiedflowtransformers,esser2021tamingtransformershighresolutionimage,rombach2022highresolutionimagesynthesislatent} and understanding \cite{liu2023llava,liu2023improvedllava,dai2023instructblip,radford2021learningtransferablevisualmodels} within a single model and have demonstrated remarkable scalability and cross-task potential \cite{wu2024vila, chameleonteam2024chameleonmixedmodalearlyfusionfoundation, wu2024janus, zhuang2025argus}. However, the training of these models is prohibitively expensive; for instance, VILA-U \cite{wu2024vila} requires approximately 20K A100 GPU hours. While many prior methods propose to reduce inference-time computation in understanding-only VLMs via token pruning or special attention masks \cite{chen2024imageworth12tokens, shang2024LLaVA-PruMerge, zhang2025avladaptiveattentionlarge,bolya2023tokenmergingvitfaster,rao2021dynamicvit,luo2023cheapquickefficientvisionlanguage,hu2024matryoshkaquerytransformerlarge}, these strategies do not directly translate to improve training-time efficiency. Furthermore, existing acceleration techniques for visual understanding do not account for the distinct structural requirements of visual generation, nor do they study the complexities inherent in unifying generative and discriminative objectives within a single VLM.

\begin{figure*}[t]
    \centering
    \includegraphics[width=\textwidth]{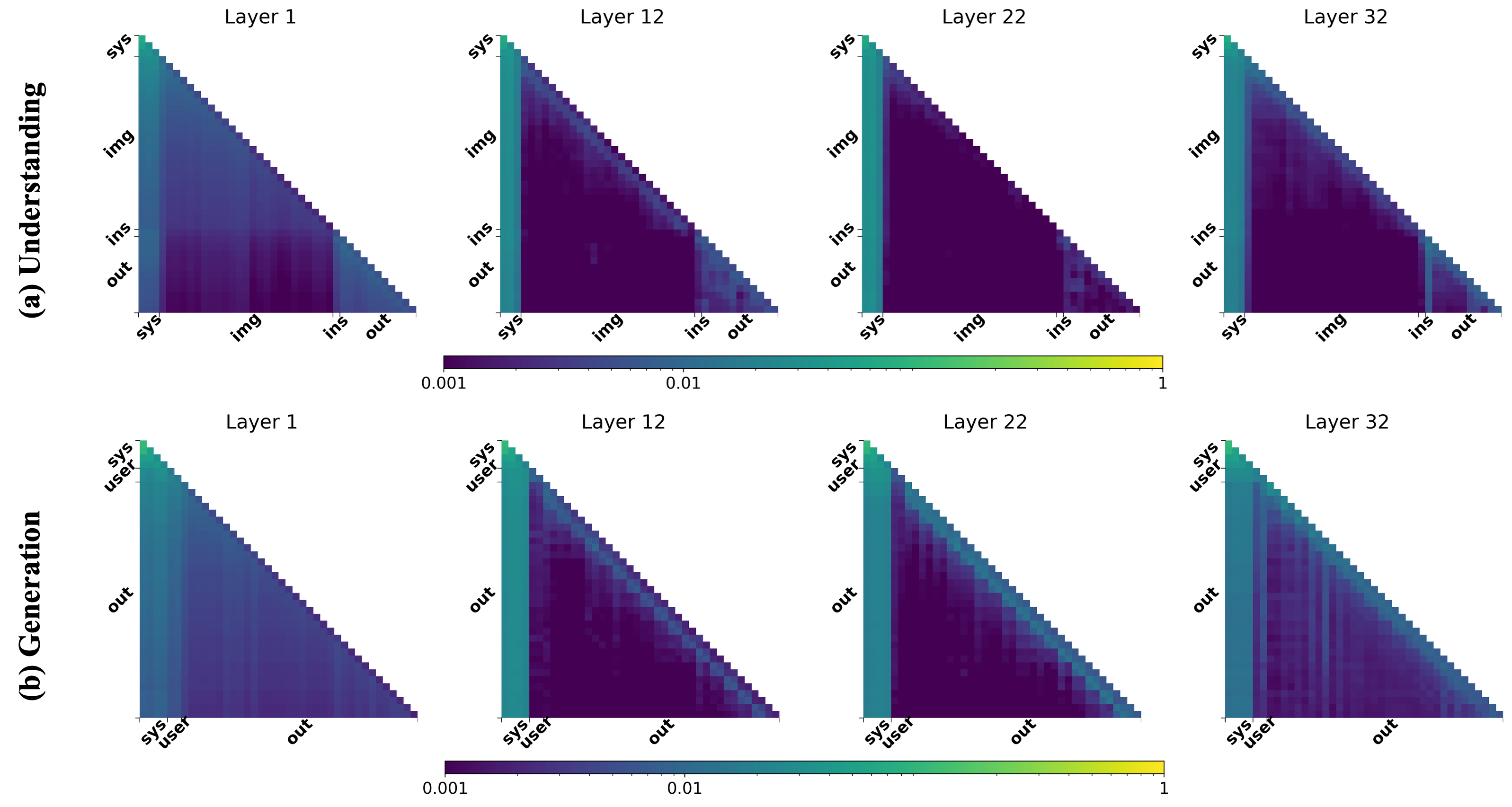}
    \caption{\textbf{Asymmetric depth-wise attention patterns in unified VLMs.}
Visualization of self-attention heatmaps across layers for understanding (a) and generation (b). Understanding exhibits strong early cross-modal interactions followed by a sharp decay in image-token attention. Generation, however, preserves substantial image-token attention throughout depth, highlighting a fundamental asymmetry in token utilization.}
    \label{fig:attn}
\vspace{-.2in}
\end{figure*}

In this paper, we investigate the feasibility and limits of accelerating the training of unified vision language models. We adopt the pure autoregressive framework as our testbed, as it represents one of the most prevalent architectures for integrating multimodal capabilities \cite{wu2024vila,wang2024emu3nexttokenpredictionneed,liu2025world,xie2024showo,yu2023scalingautoregressivemultimodalmodels,zhan2025anygptunifiedmultimodalllm,ge2023makingllamadrawseed,jin2024unifiedlanguagevisionpretrainingllm}. Through an analysis of the attention dynamics within this framework (in \Cref{fig:attn}), we reveal a critical asymmetry in task-specific redundancy: while visual understanding tasks exhibit high token redundancy in the deeper layers, visual generation depends heavily on the context of previously generated image tokens within many deep layers. Building on these insights, we develop task-specific strategies to accelerate training by selectively dropping image tokens tailored to the unique requirements of each objective.

Furthermore, we reveal a critical "synergy loss" phenomenon that occurs when task-specific token reduction methods are applied to the joint training of unified models. We find that task-specific token dropping disrupts the inherent synergy between understanding and generation by: (1) necessitating divergent sets of image-related model parameters, and (2) eliminating the mutual performance gains typically observed when both tasks are trained concurrently. Our diagnostic analysis suggests that aggressive token dropping amplifies task conflicts, offering a cautionary lesson and a new perspective for future research in efficient unified modeling. Our contributions are summarized as follows:

\begin{itemize} 
    \item \textbf{Unified Redundancy Analysis}: We characterize task-specific attention patterns in unified VLMs, identifying distinct redundancy zones. 
    \item \textbf{Task-Specific Accelerators}: We design and implement training-time acceleration for isolated tasks. 
    \item \textbf{Discovery of Synergy Loss}: We discover that task-specific optimization strategies fail in unified settings, revealing that forced token reduction disrupts mutual improvements of discriminative and generative objectives. 
    \item \textbf{Lessons for Unified Acceleration}: Our results suggest that effective acceleration methods may benefit from preserving shared cross-task structures and carefully accounting for impact on cross-task learning dynamics.

\end{itemize}

\begin{figure}[t]
    \centering
    \includegraphics[width=.9\linewidth]{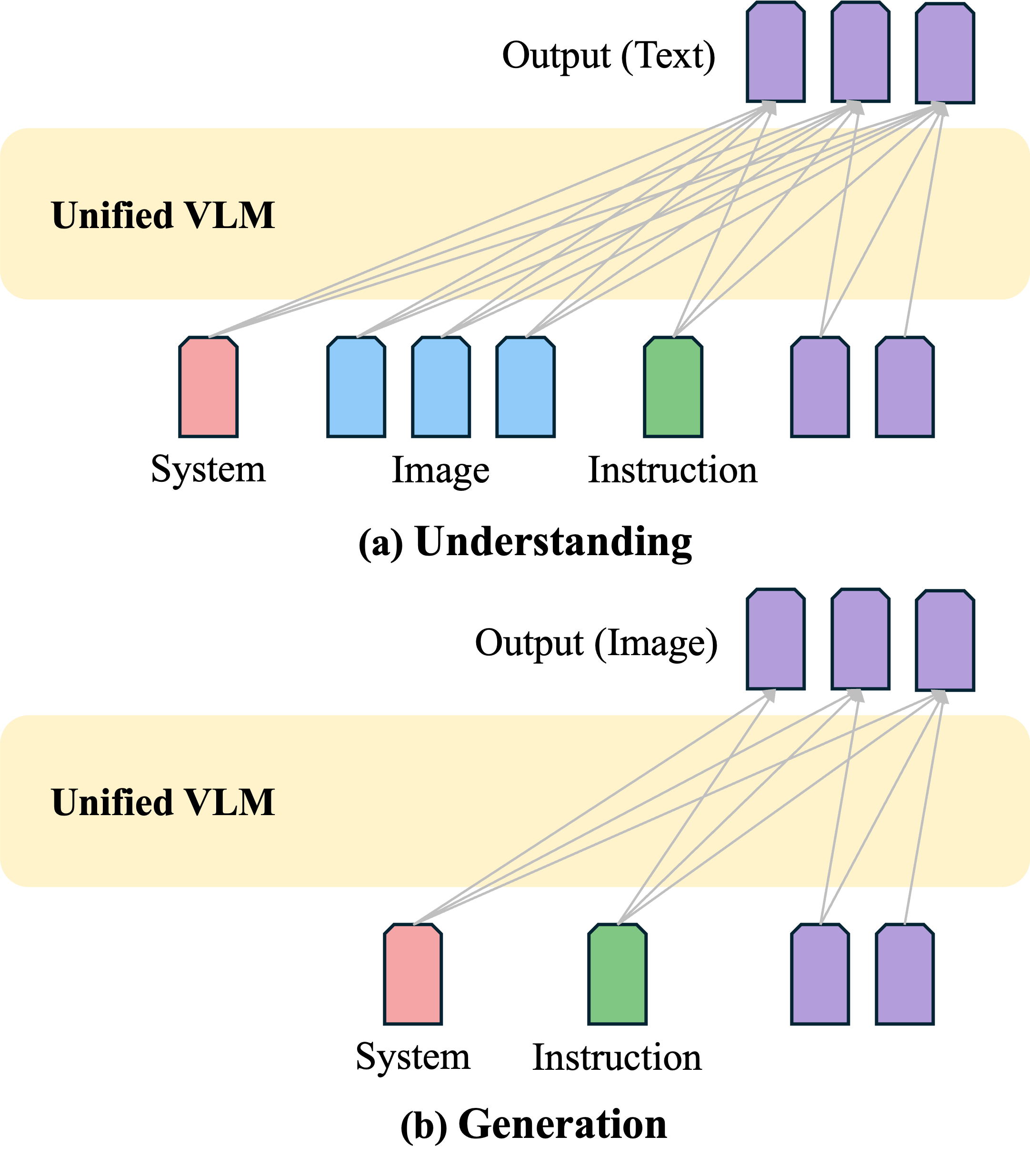}
    \vspace{-.2in}
    \caption{\textbf{Unified autoregressive VLM}.
A single Transformer backbone processes multimodal sequences under a unified next-token prediction objective. (a) In visual understanding, the model predicts text tokens conditioned on image and textual context. (b) In visual generation, the model autoregressively predicts image tokens conditioned on preceding text and image tokens.}
    \label{fig:unifed_vlm}
\vspace{-.2in}
\end{figure}

\section{Related Works}
\vspace{-.05in}

\paragraph{Unified Vision-Language Models.} Recent advancements have shifted toward unifying perception and generation within a single framework. Models like VILA-U \cite{wu2024vila}, Janus \cite{wu2024janus}, and Chameleon \cite{chameleonteam2024chameleonmixedmodalearlyfusionfoundation} utilize discrete visual tokenizers (e.g., VQVAE \cite{oord2018neuraldiscreterepresentationlearning}) to treat images as a "foreign language." While these models simplify the pipeline by using a single next-token prediction objective, their joint training is computationally demanding. Many other hybrid models that append diffusion heads \cite{rombach2022highresolutionimagesynthesislatent} to a transformer also require fine-tuning the entire backbone across multiple modalities, creating the need for efficient training.

\vspace{-.2in}
\paragraph {Efficiency in Vision-Language Models.} Efficiency research in VLMs has primarily focused on visual understanding during inference-time \cite{llavamini,shang2024LLaVA-PruMerge,chen2024imageworth12tokens,li2024tokenpackerefficientvisualprojector,hu2024matryoshkaquerytransformerlarge,luo2023cheapquickefficientvisionlanguage,lin2025boostingmultimodallargelanguage}. For instance, LLaVA-PruMerge \cite{shang2024LLaVA-PruMerge} and LLaMA-VID \cite{li2023llamavidimageworth2} reduce the number of visual tokens by identifying spatial redundancy and merging tokens. Other works explore efficient attention mechanisms or special masks to skip redundant computations during inference \cite{zhang2025avladaptiveattentionlarge, zhang2025himixreducingcomputationalcomplexity}. However, these methods are often designed for "understanding-only" tasks where the model's output is limited to text, and a complete set of image tokens is treated as input, thus having difficulty applying to visual generation, and how to reduce training-time computation remains a challenging problem.

\vspace{-.15in}
\paragraph{Token Reduction and Attention Redundancy.} The concept of "token reduction" or "pruning" originates from the Vision Transformer (ViT) and NLP literature to handle long-sequence data \cite{rao2021dynamicvit, bolya2023tokenmergingvitfaster,xiao2023streamingllm,gu2024attention}. These methods typically use attention weights or activation statistics as proxies for token importance. In the multimodal domain, recent studies have analyzed attention sinks \cite{xiao2023streamingllm} and sparsity to prune background patches. While effective for single-task models, these importance metrics are not directly transferable to unified models where tokens must serve dual roles in discriminative perception and generative synthesis.

\vspace{-.15in}
\paragraph{Multi-task Synergy in VLMs.} The relationship between understanding and generation has been a subject of ongoing debate. While some studies suggest that generative pre-training provides a stronger world model for perception \cite{wang2024emu3nexttokenpredictionneed, xie2024showo}, others have noted the difficulty of balancing these disparate objectives during joint optimization \cite{wu2024janus}. We build upon this line of inquiry by investigating how structural constraints—specifically, token dropping—affect the stability and synergy of this multi-task learning process.
\vspace{-.15in}
\section{Problem Setup}
\vspace{-.05in}
\subsection{Unified Autoregressive Vision-Language Model}

We study a unified vision-language model (VLM) that jointly performs visual understanding and visual generation within a single autoregressive Transformer backbone, following the unified next-token prediction paradigm of VILA-U (7B)~\cite{wu2024vila}. Let $x = (x_1, \ldots, x_T)$ denote text tokens and $v = (v_1, \ldots, v_M)$ denote discrete image tokens obtained from a visual tokenizer (e.g., VQ-based). We construct a multimodal sequence
\[
z = (z_1, \ldots, z_{|z|})
= (\text{system},\, x,\, v).
\]

The model, parameterized by $\theta$, is trained using autoregressive next-token prediction:
\[
P_\theta(z)
=
\prod_{t=1}^{|z|}
P_\theta(z_t \mid z_{<t}).
\]
Under this formulation, both text and image tokens are treated uniformly as discrete tokens in a single sequence, and a shared next-token objective is applied across modalities. We visualize the generation process in \Cref{fig:unifed_vlm}.

\vspace{-.05in}
\subsection{Training Objective}
\vspace{-.05in}

Unified training mixes data from visual understanding and visual generation tasks under a single objective. For a multimodal training sample $z$, the loss is defined as the negative log-likelihood:
\begin{equation}
\mathcal{L}_{\text{unified}}
=
-
\sum_{t=1}^{|z|}
\log
P_\theta(z_t \mid z_{<t}).
\end{equation}

This unified objective simultaneously optimizes: \textbf{Visual Understanding:} predicting text tokens conditioned on image and textual context; \textbf{Visual Generation:} predicting image tokens conditioned on preceding text and image tokens.

\vspace{-.05in}
\subsection{Computational Bottleneck}
\vspace{-.05in}

Let $N = |z|$ denote the total sequence length. A standard Transformer layer incurs quadratic self-attention cost: $
\text{FLOPs per layer}
\propto
N^2.
$
Since $N = T + M$, where $T$ and $M$ are the numbers of text and image tokens respectively,
\[
(T + M)^2
=
T^2 + 2TM + M^2.
\]
In unified VLM training, image tokens typically dominate the sequence ($M \gg T$), making the $M^2$ term the primary computational bottleneck.

This naturally motivates token reduction strategies that limit the effective participation of image tokens in attention. In this work, we investigate the effectiveness and limitations of \textit{token-reduction-based training acceleration} for visual understanding, visual generation, and their unification. We begin by analyzing task-specific redundancy patterns through attention statistics.

\begin{figure*}[t]
    \centering
    \includegraphics[width=\textwidth]{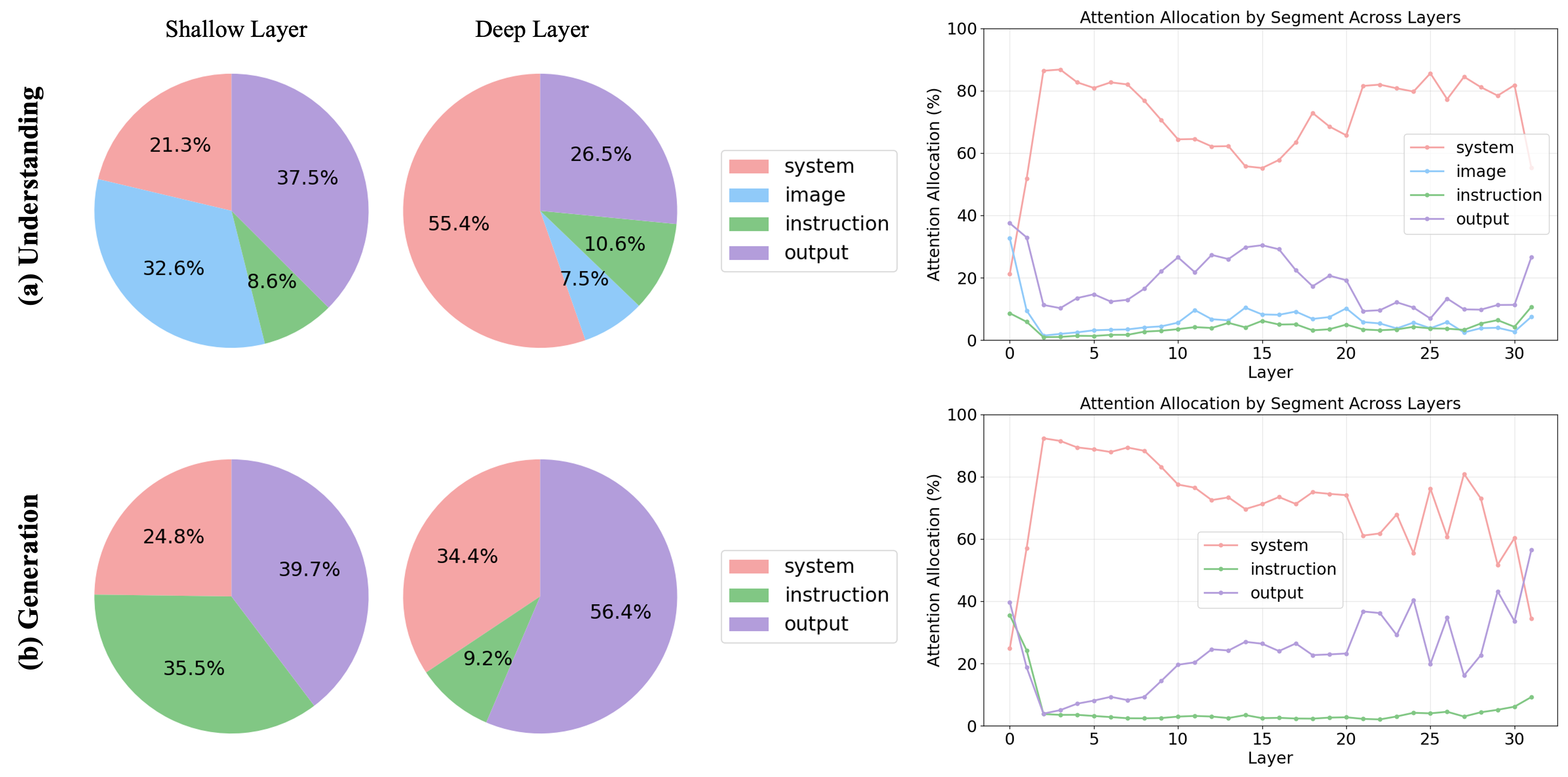}
    \caption{\textbf{Quantitative attention allocation reveals depth-dependent visual redundancy asymmetry.}
Left: Attention mass distribution over token segments (system, image, instruction, output) at representative shallow and deep layers. Right: Layerwise attention allocation across the full transformer depth. For visual understanding (top), attention to image tokens sharply decreases in deeper layers, while instruction and output tokens dominate, indicating substantial late-layer visual redundancy. In contrast, visual generation (bottom) maintains consistently high attention to image tokens across layers, with increasing allocation to output image tokens in deep layers, reflecting persistent autoregressive dependence on generated image tokens.}
    \label{fig:stats}
\vspace{-.2in}
\end{figure*}

\vspace{-.1in}
\section{Redundancy Analysis}
\vspace{-.1in}

To guide the design of our acceleration strategies, we analyze the layerwise attention behavior of a pre-trained unified VLM. Our goal is to analyze task-specific redundancy in visual tokens that inspires method design.

\vspace{-.1in}
\subsection{Analysis Setup}

\paragraph{Model and Data.}
We analyze the VILA-U model and collect attention statistics on both visual understanding (with ShareGPT-4v dataset) and visual generation (with JournyDB dataset).
For each task, we record attention maps across all transformer layers.

\vspace{-.15in}
\paragraph{Attention Allocation.}
Following prior attention decomposition analysis~\cite{chen2024imageworth12tokens},
we measure how attention mass is distributed across token segments.
Let $A^{(\ell,h)}_{i,j}$ denote the attention weight at layer $\ell$ and head $h$ from query token $i$ to key token $j$, with
\[
\sum_j A^{(\ell,h)}_{i,j} = 1.
\]

Given a partition of tokens into segments
(e.g., \texttt{system}, \texttt{image}, \texttt{instruction}, \texttt{output}),
the \emph{attention allocation} of segment $S$ at layer $\ell$ is defined as:
\begin{equation}
\alpha^{(\ell)}_{S}
=
\frac{1}{H}
\sum_{h=1}^{H}
\sum_{i}
\sum_{j \in S}
A^{(\ell,h)}_{i,j}.
\end{equation}

This metric captures the fraction of total attention mass directed to each token segment at a given layer. We use $\alpha^{(\ell)}_{S}$ to quantify redundancy patterns across depth.

\subsection{Task-Specific Attention Patterns}

We visualize (1) attention allocation (\Cref{fig:stats}) across token segments over layers and (2) attention heatmaps (\Cref{fig:attn}) at representative layers for both visual understanding (U) and visual generation (G). The results reveal a clear asymmetry in visual token redundancy patterns in different tasks.

\vspace{-.15in}
\paragraph{Visual Understanding (U).}
For perception tasks (e.g., VQA), visual tokens exhibit clear depth-dependent redundancy. 
As shown in \Cref{fig:attn} and \Cref{fig:stats}, attention rapidly shifts away from image tokens as depth increases. 
Image tokens account for roughly $\sim$30\% of attention in the first layer, but this drops below 10\% in middle and late layers. 
Instead, attention becomes dominated by instruction and output tokens, which together exceed 80\% of the total attention mass in deeper layers. Across layers, we observe a consistent transition:

\begin{itemize}
    \item \textbf{Early layers:} Strong cross-modal interactions between image and text tokens, indicating visual grounding and alignment.
    \item \textbf{Middle layers:} Attention increasingly concentrates on text tokens, with diminishing image-to-image and image-to-text interactions.
    \item \textbf{Late layers:} Attention is almost entirely confined to textual tokens, suggesting that high-level reasoning becomes predominantly linguistic.
\end{itemize}

\begin{figure*}[t]
    \centering
    \includegraphics[width=\textwidth]{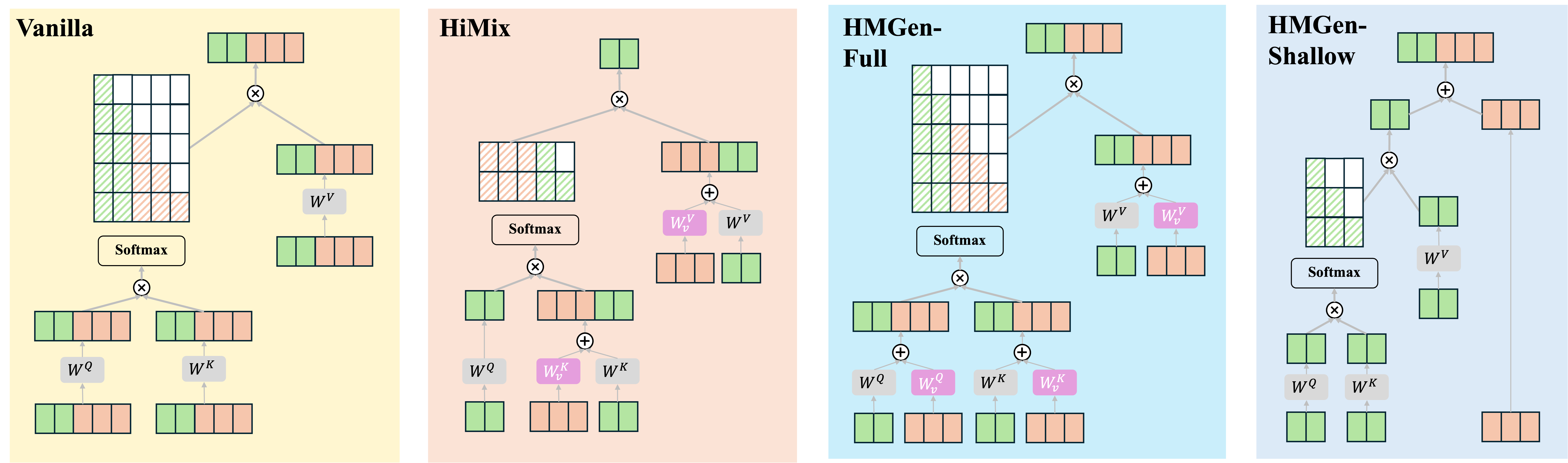}
    \caption{\textbf{Task-specific token-reduction-based acceleration mechanisms for unified VLM training.}
From left to right: (1) Vanilla Transformer layer, where both text and image tokens participate fully in self-attention and feed-forward computation. (2) HiMix (Understanding) reduce image tokens from the query stream while retaining them in key/value projections, eliminating quadratic image-to-image attention while preserving text-to-image interactions. (3) HMGen–Full layer (Generation) maintains full autoregressive attention but separates image- and text-related projections for stable hierarchical conditioning. (4) HMGen–Shallow layer (Generation) skips image-token attention and feed-forward updates, forwarding their hidden states to reduce computation while preserving autoregressive structure.}
    \label{fig:attention_method}
\vspace{-.2in}
\end{figure*}

\vspace{-.15in}
\paragraph{Visual Generation (G).}
In contrast to understanding, image generation exhibits a persistent and structured dependence on image tokens. 
Output (image) tokens receive a substantial fraction of attention across early and late layers, typically ranging from 30\% to 60\%. Unlike the rapid decay observed in understanding tasks, attention to image tokens exhibits a consistent increase of attention allocation in deeper layers. Across depth, the attention pattern follows a hierarchical structure:
\begin{itemize}
    \item \textbf{Early layers:} Broad attention over both textual prompts and previously generated image tokens, establishing global conditioning.
    \item \textbf{Middle layers:} Attention concentrates on recent image tokens and specific prefix positions, reflecting localized autoregressive dependencies.
    \item \textbf{Late layers:} Image-token attention becomes increasingly significant, ensuring consistency in token prediction.
\end{itemize}

\vspace{-.15in}
\paragraph{Robustness Across Scales.}
We repeat the same analysis on a smaller-scale VILA-U model trained by ourselves (LLaMA-3-3B backbone \cite{dubey2024llama3}). The qualitative and quantitative patterns remain consistent: late-layer visual redundancy emerges for understanding, while generation preserves significant image-token attention. This suggests the observed asymmetry is not scale-specific.

\subsection{Implications for Acceleration}

This implies token reduction must be task-aware:

\begin{enumerate}
    \item \textbf{Understanding:} Visual tokens are redundant after the first few layers. It is possible to reduce image tokens in some way and significantly reduce quadratic attention cost with minimal performance impact.
    
    \item \textbf{Generation:} Visual tokens are autoregressively generated during inference, and removing training computation on them must still enable the same next-token prediction for inference. Deep layers in visual generation also have limited bandwidth to reduce image token-related computation.
\end{enumerate}

Therefore, a unified model cannot rely on a single token-dropping rule. The structural roles of visual tokens differ fundamentally between discriminative and generative objectives. We introduce the task-specific training acceleration methods below.

\vspace{-.1in}
\section{Proposed Task-Specific Accelerators} 
\vspace{-.1in}

\begin{table*}[t]
\centering
\caption{\textbf{HiMix for visual understanding.}
Performance and computational cost comparison between the understanding-only VILA-U baseline and HiMix. HiMix reduces training FLOPs to 0.24× (76\% reduction) by removing image-token queries, while incurring only moderate performance degradation across GQA, MME, POPE, and SeedBench benchmarks. The relatively small accuracy drop compared to the substantial computational savings confirms significant late-layer visual redundancy in understanding tasks.}
\resizebox{\linewidth}{!}{
\begin{tabular}{lccccccccc}
\toprule
Method 
& GQA 
& MME-C 
& MME-P 
& POPE-A 
& POPE-P 
& POPE-R 
& POPE-F1 
& SeedBench-Img
& FLOPs \\
\midrule
VILA-U (U-only) & 52.86 & 258.21 & 1054.88 & 81.30 & 84.67 & 74.76 & 79.40 & 46.05 & 1$\times$ \\
HiMix (U-only)  & 49.92 & 224.64 & 983.30 & 78.56 & 78.03 & 79.49 & 78.75 & 40.88 & 0.24$\times$ \\
\bottomrule
\end{tabular}
}
\label{tab:himix_only}
\vspace{-.2in}
\end{table*}

Motivated by the task-specific redundancy revealed in Sec.~4, we investigate whether token-reduction-based acceleration can be done separately for visual understanding and generation. We first evaluate these strategies in isolation before analyzing their behavior under unified training.

\subsection{Understanding (U)} 

\paragraph{Method.}
We adopt HiMix~\cite{zhang2025himixreducingcomputationalcomplexity} as the baseline accelerator for visual understanding. Unlike token merging/dropping given complete image token sets based on inference-time analysis, HiMix modifies the attention computation in a manner compatible with both training and inference, making it suitable for unified autoregressive VLMs. The key idea is to \textit{reduce tokens in queries}.

Specifically, as illustrated in \Cref{fig:attention_method}, image tokens are removed from the \textit{query} projections while retained in the \textit{key} and \textit{value} projections. This eliminates quadratic image-to-image attention while preserving text-to-image interactions. As shown in Sec.~4, visual tokens become increasingly redundant in deeper layers for understanding tasks; thus, removing them from queries reduces computation with minimal impact on prediction. Moreover, noticing that the final output of each layer only includes text tokens as image tokens are removed from queries, this strategy requires the input original image tokens to each layer of the transformer.

\vspace{-.15in}
\paragraph{Theoretical Efficiency.}
For a sequence with $T$ text tokens and $M$ image tokens (total length $T+M$) and hidden size $d$, the per-layer complexity of a vanilla Transformer can be decomposed into:
(i) self-attention, dominated by the $QK^\top$ operation, $\mathcal{O}((T+M)^2 d)$; and
(ii) the feed-forward network (two linear layers), $\mathcal{O}(8(T+M)d^2)$.
Thus,
\[
\text{Cost}_{\text{base}} \;=\; \mathcal{O}\!\left((T+M)^2 d \;+\; 8(T+M)d^2\right).
\]
With HiMix, image tokens are removed from the \emph{query} stream, so attention is computed only for $T$ text queries over $(T+M)$ keys/values, reducing the attention term to $\mathcal{O}(T(T+M)d)$ while keeping the FFN term unchanged:
\[
\text{Cost}_{\text{HiMix}} \;=\; \mathcal{O}\!\left(T(T+M)d \;+\; 8Td^2\right).
\]
When $M \gg T$, the dominant $\mathcal{O}(M^2 d)$ attention term is removed. In practice, this leads to substantial FLOPs reduction while preserving the cross-modal interactions necessary for visual understanding.

\vspace{-.15in}
\paragraph{Experimental Results.}
We evaluate HiMix in an understanding-only setting of VILA-U (LLaMA-3-3B backbone \cite{dubey2024llama3,touvron2023llamaopenefficientfoundation}), with the ShareGPT-4v dataset \cite{chen2024sharegpt4v}. We follow VILA-U to conduct pretraining and finetuning each for one epoch, and evaluate on several visual understanding benchmarks \cite{hudson2019gqanewdatasetrealworld,fu2025mme,li2023evaluatingobjecthallucinationlarge,li2023seedbenchbenchmarkingmultimodalllms}. Results are in Table~\ref{tab:himix_only}. 

HiMix reduces FLOPs to $0.24\times$ of the baseline, corresponding to a 76\% reduction in computation. Despite this substantial saving, performance degradation remains moderate. For example, GQA accuracy decreases from $52.86$ to $49.92$, while POPE F1 drops slightly from $79.40$ to $78.75$. Notably, the performance drop is significantly smaller than the reduction in computational cost, indicating substantial redundancy in late-layer visual processing for understanding tasks. Overall, these results confirm that visual understanding exhibits considerable late-layer image redundancy. Structured removal of image-token queries yields large efficiency gains while largely preserving cross-modal reasoning capability.

\vspace{-.1in}
\subsection{Generation (G)}
\paragraph{Design Constraints from Autoregressive Image Generation.}

Unlike visual understanding, visual generation follows a strict autoregressive process: each predicted image token is appended to the sequence and must serve as a valid query for predicting subsequent tokens. Therefore, image tokens \emph{must remain in the query stream}. Removing them from queries would break the autoregressive dependency chain and make inference inconsistent with training. This constraint fundamentally differentiates generation from understanding and prevents directly applying HiMix-style query removal. 

One might instead consider removing image tokens from key/value projections while keeping them in queries. Although this reduces part of the attention computation, two major issues arise.
\textbf{(1) Limited FLOPs Reduction.}  
Even if image-to-image attention is partially suppressed, the feed-forward network (FFN) still processes all image tokens. When $M \gg T$, the dominant $\mathcal{O}(8 M d^2)$ FFN term remains intact, resulting in minimal overall computational savings.
\textbf{(2) Severe Performance Degradation.}  
Image generation exhibits persistent image-token dependence across depth (Sec.~4). Suppressing key/value participation disrupts hierarchical autoregressive conditioning, leading to substantial quality degradation in practice. Empirically, we observe that this naive modification yields both limited efficiency gains and large drops in generative performance. For example, applying this design to one middle layer leads to a significant drop (-3.52) on MJHQ-30K \cite{li2024playgroundv25insightsenhancing}.

\vspace{-.15in}
\paragraph{HMGen: Hierarchical Mixture for Generation.}
Motivated by the hierarchical attention structure observed in Sec.~4, we instead propose \textbf{HMGen}, which is composed of two kinds of layers illustrated in Figure~\ref{fig:attention_method}. HMGen preserves the autoregressive structure with image in query \textit{from model level} while reducing tokens in specific layers.

We introduce $K$ designated \emph{shallow layers} in the middle portion of the transformer (out of $L$ total layers) while other layers remain as \emph{full layers}. This is because early full layers are required to preserve global conditioning, while late full layers are required to ensure high-fidelity final token prediction. We empirically find that \textit{alternating shallow and full layers in the middle layers} yields the best trade-off between efficiency and generation quality. 

In \textit{shallow layers}, image-token attention computation is skipped, and the feed-forward network is applied only to text tokens. The image-token hidden states are directly forwarded from the previous layer to the next without participating in self-attention or FFN updates.

In \textit{full layers}, we further introduce separate projection parameters for image and text tokens. Although the backbone remains unified, decoupling image-related projections stabilizes training and improves generation quality. This separation allows image-token representations to maintain dedicated pathways even when their participation in attention is selectively reduced. Empirically, we observe improved performance compared to fully shared parameterization under the same FLOPs budget.

\begin{figure}[t]
    \centering
    \includegraphics[width=\linewidth]{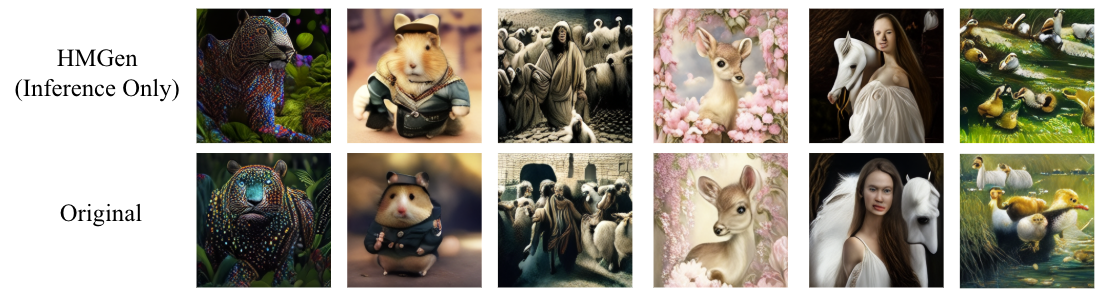}
    \caption{\textbf{Inference-time-only HMGen.}
Qualitative comparison between the original model (bottom) and inference-time-only HMGen (top), where image-token computation in shallow layers is skipped without retraining. Visual quality and semantic consistency are largely preserved despite reduced computation.}
    \label{fig:inf_only}
\vspace{-.2in}
\end{figure}

\vspace{-.15in}
\paragraph{Theoretical Efficiency.}
HMGen maintains the autoregressive dependency chain while reducing computation in $K$ designated middle ``shallow'' layers (out of $L$ total) by skipping image-token attention/MLP computation and forwarding their hidden states. Using the same decomposition as above, a vanilla layer costs
\[
\text{Cost}_{\text{base}} \;=\; \mathcal{O}\!\left((T+M)^2 d \;+\; 8(T+M)d^2\right).
\]
In a shallow layer, attention is computed only for $T$ text queries, giving $\mathcal{O}(T^2d)$, and the FFN is applied only to text tokens, giving $\mathcal{O}(8Td^2)$:
\[
\text{Cost}_{\text{shallow}} \;=\; \mathcal{O}\!\left(T^2d \;+\; 8T d^2\right).
\]
The total complexity across $L$ layers is therefore
\begin{align*}
\text{Cost}_{\text{HMGen}}
=&\; (L-K)\,\mathcal{O}\!\left((T+M)^2 d + 8(T+M)d^2\right) \nonumber \\
&+ K\,\mathcal{O}\!\left(T^2 d + 8T d^2\right).
\end{align*}
When $M \gg T$, each shallow layer removes the dominant image-related costs in both attention and FFN, i.e., the $\mathcal{O}(M^2 d)$ and $\mathcal{O}(8Md^2)$ terms. Consequently, the overall FLOPs reduction scales with the fraction of layers made shallow ($K/L$), and is upper-bounded by the compute in the remaining $(L-K)$ full layers. In the idealized regime where shallow layers contribute negligible cost relative to full layers, the relative cost approaches $1-K/L$, yielding an approximate speedup of $1/(1-K/L)$.

\vspace{-.15in}
\paragraph{Experimental Results.}
We first evaluate HMGen in a generation-only setting of VILA-U using the JourneyDB \cite{sun2023journeydb} dataset, and evaluate visual generation on MJHQ-30K \cite{li2024playgroundv25insightsenhancing}. Quantitative results are shown in Table~\ref{tab:hmgen_only}, and qualitative inference-time only results (without training, just directly skipping image-related computations in middle layers) are visualized in Figure~\ref{fig:inf_only}.

\noindent \textbf{(1) Inference-Time Applicability.}
Figure~\ref{fig:inf_only} demonstrates that HMGen can be directly applied at inference time without architectural modification. By design, shallow layers preserve the autoregressive query structure, allowing image tokens to be appended and used as subsequent queries during generation. This confirms that HMGen is not merely a training-time approximation but a structurally consistent acceleration mechanism.

\noindent \textbf{(2) Reasonable FLOPs Reduction.}
As shown in Table~\ref{tab:hmgen_only}, introducing $K$ shallow layers yields significant computational savings. With $K=3$, FLOPs are reduced to $0.85\times$ of the baseline, and with $K=5$, to $0.75\times$. Since each shallow layer removes both the dominant $\mathcal{O}(M^2 d)$ attention term and the $\mathcal{O}(8 M d^2)$ FFN term, the efficiency gain scales approximately with the fraction of shallow layers.

\noindent \textbf{(3) Improved Generation Quality.}
Notably, HMGen achieves substantially better MJHQ-30K scores compared to the generation-only VILA-U baseline (17.45 $\rightarrow$ 12.16 with $K=3$). This improvement arises from our separation of image and text projection parameters within the full layers. By decoupling image-specific transformations, the model maintains more stable hierarchical image representations even when computation is selectively reduced. 

Overall, HMGen not only reduces computation but also enhances generative quality, demonstrating that hierarchical, structure-aware acceleration is better aligned with the intrinsic dependencies of visual generation.

\begin{table}[t]
\centering
\caption{\textbf{HMGen for visual generation.}
Introducing shallow layers reduces FLOPs (to 0.85× and 0.75×) while improving generative quality compared to the VILA-U baseline, demonstrating efficient and structure-aware acceleration.}
\resizebox{\linewidth}{!}{
\begin{tabular}{lccc}
\toprule
Method 
& \#Shallow Layers 
& MJHQ-30K 
& FLOPs \\
\midrule
VILA-U (G-only) & 0 & 17.45 & 1$\times$ \\
HMGen & 3 & \textbf{12.16} & 0.85x \\
HMGen & 5 & \textbf{12.55} & 0.75x \\
\bottomrule
\end{tabular}
}
\label{tab:hmgen_only}
\vspace{-.2in}
\end{table}

\begin{table*}[t]
\centering
\setlength{\tabcolsep}{3.5pt}
\caption{
\textbf{Unified training performance and efficiency.}
The unified baseline improves both understanding (e.g., GQA 0.5600 vs.\ 0.5286 U-only) 
and generation (MJHQ 15.78 vs.\ 17.45 G-only), demonstrating positive cross-task synergy. 
However, combining HiMix and HMGen under joint training substantially reduces FLOPs 
(0.55--0.56$\times$) but degrades performance on both objectives 
(e.g., GQA drops to 0.4705/0.3300 and MJHQ worsens to 14.54/12.53), 
indicating that task-specific token reduction disrupts mutual gains.
}
\vspace{-.1in}
\resizebox{\linewidth}{!}{
\begin{tabular}{lcccccccccc}
\toprule
Method 
& GQA 
& MME-C 
& MME-P 
& POPE-A 
& POPE-P 
& POPE-R 
& POPE-F1 
& SeedBench-Img
& MJHQ 
& FLOPs \\
\midrule

VILA-U (U-only) 
& 52.86 & 258.21 & 1054.88 & 81.30 & 84.67 & 74.76 & 79.40 & 46.05 & -- & 1$\times$ \\

VILA-U (G-only) 
& -- & -- & -- & -- & -- & -- & -- & -- & 17.45 & 1$\times$ \\

VILA-U (Unified) 
& 56.00 & 250.00 & 1135.91 & 83.37 & 87.95 & 77.33 & 82.30 & 47.88 & 15.78 & 1$\times$ \\

HiMix (U-only)            
& 49.92 & 224.64 & 983.30 & 78.56 & 78.03 & 79.49 & 78.75 & 40.88 & -- & 0.24$\times$ \\

HMGen (G-only)            
& -- & -- & -- & -- & -- & -- & -- & -- & 12.16 & 0.85$\times$ \\

HiMix-HMGen\\(\textit{Share All Params})     
& 33.00 & 233.21 & 662.26 & 60.84 & 57.66 & 81.64 & 67.59 & 31.38 & 12.53 & 0.56$\times$ \\

HiMix-HMGen\\(\textit{Share Partial Params}) 
& 47.05 & 255.00 & 847.82 & 76.43 & 76.10 & 77.10 & 76.58 & 34.50 & 14.54 & 0.55$\times$ \\

\bottomrule
\end{tabular}
}
\label{tab:unified}
\vspace{-.2in}
\end{table*}

\vspace{-.1in}
\section{The Limits of Unified Efficiency}
\vspace{-.1in}

While the task-specific token-reduction-based accelerators in Sec.~4 demonstrate substantial efficiency gains when applied to understanding or generation in isolation, our primary objective is to evaluate their behavior under unified training. In unified VLMs, both objectives are optimized jointly under a shared backbone, and improvements in one task often influence the other through shared representations. Efficiency modifications may therefore interact with cross-task learning dynamics in subtle ways. In this section, we examine whether task-specific acceleration strategies remain effective in a unified setting, and identify structural barriers that emerge during joint optimization.

\vspace{-.05in}
\subsection{Synergy Breakage: The Cost of Efficiency}
\vspace{-.05in}

\paragraph{Positive Cross-Task Synergy Baseline}

We first examine the unified baseline without token reduction. From Table~\ref{tab:unified}, joint training improves both tasks relative to their single-task counterparts. \textbf{Understanding improves under unified training:}
GQA increases from 52.86 (U-only) to 56 (Unified), POPE F1 from 79.40 to 82.3, and SeedBench from 46.05 to 47.88. \textbf{Generation also improves:}
MJHQ-30K improves from 17.45 (G-only) to 15.78 (Unified), indicating better generative quality.
Formally, let performance on understanding and generation be $\mathcal{U}(\theta)$ and $\mathcal{G}(\theta)$. For the unified baseline,
$$
\mathcal{U}(\theta_{\text{unified}}) > \mathcal{U}(\theta_{\text{U-only}}),
\quad
\mathcal{G}(\theta_{\text{unified}}) > \mathcal{G}(\theta_{\text{G-only}}).
$$
This mutual improvement confirms the presence of positive cross-task transfer, which motivates unified modeling.

\begin{figure}[t]
    \centering
    \includegraphics[width=\linewidth]{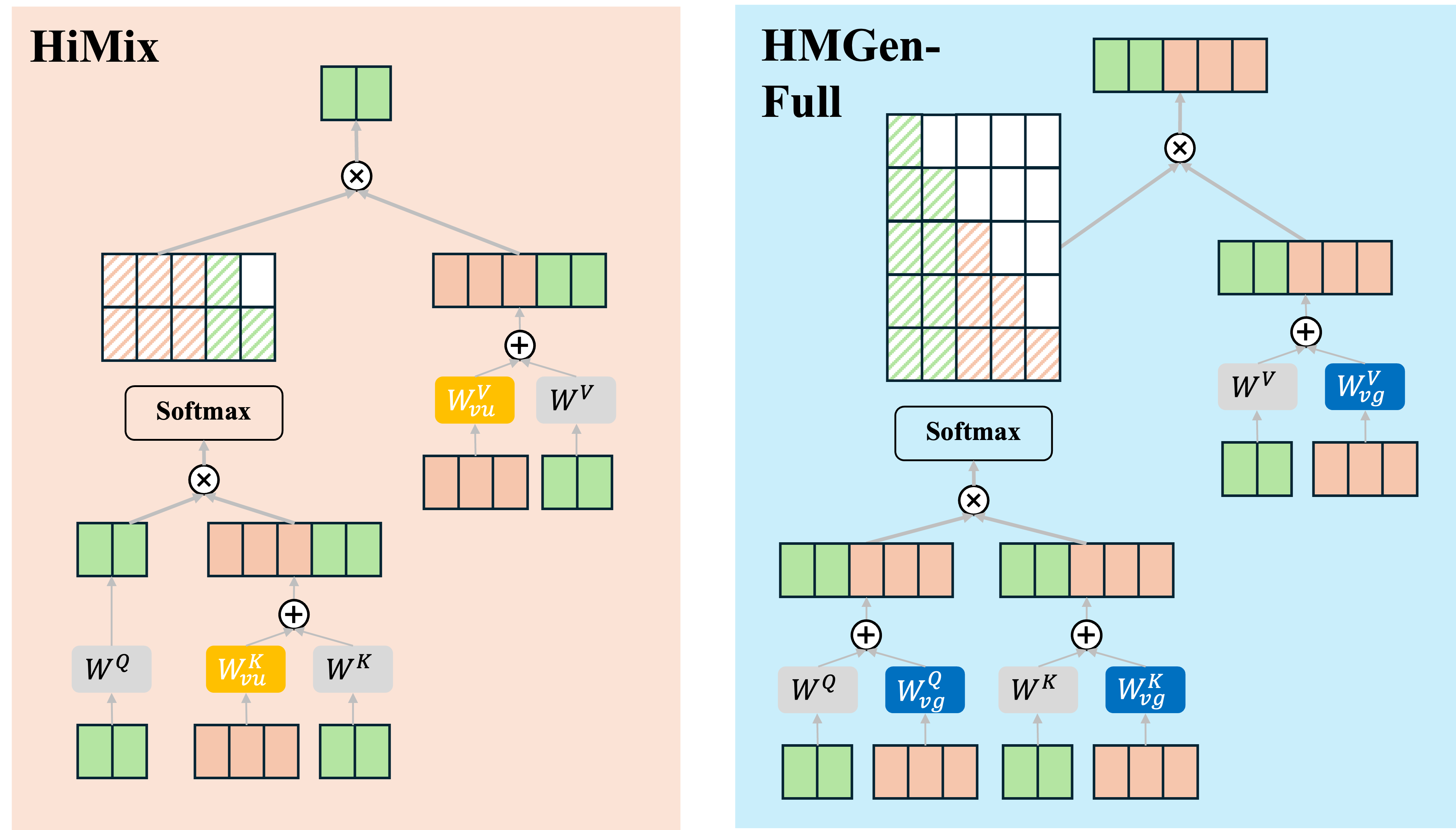}
    \caption{
\textbf{Separate image projection strategy.}
To reduce interference between task-specific routing, we decouple image-related projection parameters (e.g., $W^Q_v$, $W^K_v$, $W^V_v$) for HiMix (highlighted by yellow) and HMGen-Full (highlighted by blue), while keeping the backbone shared. 
This design aims to stabilize hierarchical image representations when token participation differs across tasks.
}
    \label{fig:separate_param}
\vspace{-.2in}
\end{figure}

\vspace{-.15in}
\paragraph{Severe Collapse with Fully Shared Parameters in HiMix-HMGen.}
The row \emph{HiMix--HMGen (Share All)} in Table~\ref{tab:unified} shows substantial degradation than HiMix (U-only) and HMGen (g-only):
GQA drops from 56 to 33, POPE F1 from 82.3 to 67.59, SeedBench from 47.88 to 31.38, while MJHQ-30K worsens from 12.16 to 12.53.
Although FLOPs are reduced to $0.56\times$, both tasks suffer significant performance collapse. Notably, unified performance becomes worse than the single-task baseline in some metrics, indicating negative transfer. Thus, naively combining task-specific accelerators destroys cross-task synergy.


\subsection{Separate Image Projection Strategy}

To mitigate this issue, we introduce partial decoupling of image-related projections, as illustrated in Figure~\ref{fig:separate_param}. Instead of fully shared projections, we decompose:
$$
W^Q_v = \{W^Q_{vu}, W^Q_{vx}\},
$$
and similarly for $W^K_v$ and $W^V_v$. This creates semi-independent image pathways while preserving the unified backbone. From Table~\ref{tab:unified}, \emph{HiMix--HMGen (Share Partial)} improves over the fully shared variant significantly:
GQA increases from 33 to 47.05, POPE F1 from 67.59 to 76.58, and MJHQ-30K from 12.53 to 14.54.
However, performance still falls short of the unified baseline per understanding, while better than the unified baseline on generation (56 GQA and 15.78 MJHQ-30K), indicating that parameter separation partially restores synergy.

\subsection{Structural Drivers of Synergy Loss}
We discuss possible drivers of the observed synergy breakage below to guide future investigation. Unified training implicitly assumes a shared latent space $
\phi(z;\theta),$
where discriminative and generative signals co-shape representations. Task-specific token dropping changes which tokens participate in attention and which parameters receive gradients. Consequently, gradients
$
\nabla_\theta \mathcal{L}_U\text{ and }
\nabla_\theta \mathcal{L}_G
$
are computed under incompatible masking operators, leading to potentially fragmented optimization dynamics. This hypothesis is also supported by the separate image projection strategy.

\vspace{-.15in}
\paragraph{Key Takeaway.}
Table~\ref{tab:unified} reveals a consistent pattern:
the unified baseline exhibits positive cross-task transfer, whereas task-specific token reduction eliminates or reverses these gains.
Efficiency improvements achieved in isolation do not compose under unified optimization. Effective unified acceleration must therefore preserve shared computational pathways that enable cross-task representation alignment, rather than simply aggregating task-optimal pruning strategies.





\vspace{-.05in}
\section{Conclusion}

We investigate the feasibility and limits of token-reduction-based acceleration for unified vision-language models and identify a fundamental asymmetry in visual token usage: visual understanding exhibits substantial late-layer redundancy, whereas visual generation maintains persistent image-token dependence across depth. Based on this insight, we design task-specific accelerators that achieve significant efficiency gains in isolated settings; however, when combined under unified training, they induce a consistent \emph{synergy loss}, as task-specific token dropping leads to divergent parameter usage and removes the mutual performance gains typically observed in joint optimization. Our findings suggest that efficient unified modeling requires preserving shared computational pathways that enable cross-task representation alignment, rather than simply aggregating task-specific strategies.

\clearpage
{
    \small
    \bibliographystyle{ieeenat_fullname}
    \bibliography{main}

@inproceedings{
    dai2023instructblip,
    title={Instruct{BLIP}: Towards General-purpose Vision-Language Models with Instruction Tuning},
    author={Wenliang Dai and Junnan Li and Dongxu Li and Anthony Tiong and Junqi Zhao and Weisheng Wang and Boyang Li and Pascale Fung and Steven Hoi},
    booktitle={Thirty-seventh Conference on Neural Information Processing Systems},
    year={2023},
    url={https://openreview.net/forum?id=vvoWPYqZJA}
}

@article{wu2024vila,
 title={Vila-u: a unified foundation model integrating visual understanding and generation},
 author={Wu, Yecheng and Zhang, Zhuoyang and Chen, Junyu and Tang, Haotian and Li, Dacheng and Fang, Yunhao and Zhu, Ligeng and Xie, Enze and Yin, Hongxu and Yi, Li and others},
 journal={arXiv preprint arXiv:2409.04429},
 year={2024}
}

@misc{chameleonteam2024chameleonmixedmodalearlyfusionfoundation,
      title={Chameleon: Mixed-Modal Early-Fusion Foundation Models}, 
      author={Chameleon Team},
      year={2024},
      eprint={2405.09818},
      archivePrefix={arXiv},
      primaryClass={cs.CL},
      url={https://arxiv.org/abs/2405.09818}, 
}

@article{wu2024janus,
  title={Janus: Decoupling visual encoding for unified multimodal understanding and generation},
  author={Wu, Chengyue and Chen, Xiaokang and Wu, Zhiyu and Ma, Yiyang and Liu, Xingchao and Pan, Zizheng and Liu, Wen and Xie, Zhenda and Yu, Xingkai and Ruan, Chong and others},
  journal={arXiv preprint arXiv:2410.13848},
  year={2024}
}

@misc{touvron2023llamaopenefficientfoundation,
      title={LLaMA: Open and Efficient Foundation Language Models}, 
      author={Hugo Touvron and Thibaut Lavril and Gautier Izacard and Xavier Martinet and Marie-Anne Lachaux and Timothée Lacroix and Baptiste Rozière and Naman Goyal and Eric Hambro and Faisal Azhar and Aurelien Rodriguez and Armand Joulin and Edouard Grave and Guillaume Lample},
      year={2023},
      eprint={2302.13971},
      archivePrefix={arXiv},
      primaryClass={cs.CL},
      url={https://arxiv.org/abs/2302.13971}, 
}

@article{xie2024showo,
  title={Show-o: One Single Transformer to Unify Multimodal Understanding and Generation},
  author={Xie, Jinheng and Mao, Weijia and Bai, Zechen and Zhang, David Junhao and Wang, Weihao and Lin, Kevin Qinghong and Gu, Yuchao and Chen, Zhijie and Yang, Zhenheng and Shou, Mike Zheng},
  journal={arXiv preprint arXiv:2408.12528},
  year={2024}
}

@misc{ho2020denoisingdiffusionprobabilisticmodels,
      title={Denoising Diffusion Probabilistic Models}, 
      author={Jonathan Ho and Ajay Jain and Pieter Abbeel},
      year={2020},
      eprint={2006.11239},
      archivePrefix={arXiv},
      primaryClass={cs.LG},
      url={https://arxiv.org/abs/2006.11239}, 
}

@misc{rombach2022highresolutionimagesynthesislatent,
      title={High-Resolution Image Synthesis with Latent Diffusion Models}, 
      author={Robin Rombach and Andreas Blattmann and Dominik Lorenz and Patrick Esser and Björn Ommer},
      year={2022},
      eprint={2112.10752},
      archivePrefix={arXiv},
      primaryClass={cs.CV},
      url={https://arxiv.org/abs/2112.10752}, 
}

@misc{esser2024scalingrectifiedflowtransformers,
      title={Scaling Rectified Flow Transformers for High-Resolution Image Synthesis}, 
      author={Patrick Esser and Sumith Kulal and Andreas Blattmann and Rahim Entezari and Jonas Müller and Harry Saini and Yam Levi and Dominik Lorenz and Axel Sauer and Frederic Boesel and Dustin Podell and Tim Dockhorn and Zion English and Kyle Lacey and Alex Goodwin and Yannik Marek and Robin Rombach},
      year={2024},
      eprint={2403.03206},
      archivePrefix={arXiv},
      primaryClass={cs.CV},
      url={https://arxiv.org/abs/2403.03206}, 
}

@misc{tian2024visualautoregressivemodelingscalable,
      title={Visual Autoregressive Modeling: Scalable Image Generation via Next-Scale Prediction}, 
      author={Keyu Tian and Yi Jiang and Zehuan Yuan and Bingyue Peng and Liwei Wang},
      year={2024},
      eprint={2404.02905},
      archivePrefix={arXiv},
      primaryClass={cs.CV},
      url={https://arxiv.org/abs/2404.02905}, 
}

@inproceedings{liu2023llava,
    author      = {Liu, Haotian and Li, Chunyuan and Wu, Qingyang and Lee, Yong Jae},
    title       = {Visual Instruction Tuning},
    booktitle   = {NeurIPS},
    year        = {2023}
  }

@misc{liu2023improvedllava,
      author={Liu, Haotian and Li, Chunyuan and Li, Yuheng and Lee, Yong Jae},
      title={Improved Baselines with Visual Instruction Tuning}, 
      publisher={arXiv:2310.03744},
      year={2023},
}

@misc{radford2021learningtransferablevisualmodels,
      title={Learning Transferable Visual Models From Natural Language Supervision}, 
      author={Alec Radford and Jong Wook Kim and Chris Hallacy and Aditya Ramesh and Gabriel Goh and Sandhini Agarwal and Girish Sastry and Amanda Askell and Pamela Mishkin and Jack Clark and Gretchen Krueger and Ilya Sutskever},
      year={2021},
      eprint={2103.00020},
      archivePrefix={arXiv},
      primaryClass={cs.CV},
      url={https://arxiv.org/abs/2103.00020}, 
}

@misc{ma2024janusflowharmonizingautoregressionrectified,
      title={JanusFlow: Harmonizing Autoregression and Rectified Flow for Unified Multimodal Understanding and Generation}, 
      author={Yiyang Ma and Xingchao Liu and Xiaokang Chen and Wen Liu and Chengyue Wu and Zhiyu Wu and Zizheng Pan and Zhenda Xie and Haowei Zhang and Xingkai yu and Liang Zhao and Yisong Wang and Jiaying Liu and Chong Ruan},
      year={2024},
      eprint={2411.07975},
      archivePrefix={arXiv},
      primaryClass={cs.CV},
      url={https://arxiv.org/abs/2411.07975}, 
}

@article{liquid,
      title={Liquid: Language Models are Scalable and Unified Multi-modal Generators},
      author={Wu, Junfeng and Jiang, Yi and Ma, Chuofan and Liu, Yuliang and Zhao, Hengshuang and Yuan, Zehuan and Bai, Song and Bai, Xiang},
      journal={arXiv preprint arXiv:2412.04332},
      year={2024}
}

@misc{wang2024emu3nexttokenpredictionneed,
      title={Emu3: Next-Token Prediction is All You Need}, 
      author={Xinlong Wang and Xiaosong Zhang and Zhengxiong Luo and Quan Sun and Yufeng Cui and Jinsheng Wang and Fan Zhang and Yueze Wang and Zhen Li and Qiying Yu and Yingli Zhao and Yulong Ao and Xuebin Min and Tao Li and Boya Wu and Bo Zhao and Bowen Zhang and Liangdong Wang and Guang Liu and Zheqi He and Xi Yang and Jingjing Liu and Yonghua Lin and Tiejun Huang and Zhongyuan Wang},
      year={2024},
      eprint={2409.18869},
      archivePrefix={arXiv},
      primaryClass={cs.CV},
      url={https://arxiv.org/abs/2409.18869}, 
}

@misc{oord2018neuraldiscreterepresentationlearning,
      title={Neural Discrete Representation Learning}, 
      author={Aaron van den Oord and Oriol Vinyals and Koray Kavukcuoglu},
      year={2018},
      eprint={1711.00937},
      archivePrefix={arXiv},
      primaryClass={cs.LG},
      url={https://arxiv.org/abs/1711.00937}, 
}

@misc{esser2021tamingtransformershighresolutionimage,
      title={Taming Transformers for High-Resolution Image Synthesis}, 
      author={Patrick Esser and Robin Rombach and Björn Ommer},
      year={2021},
      eprint={2012.09841},
      archivePrefix={arXiv},
      primaryClass={cs.CV},
      url={https://arxiv.org/abs/2012.09841}, 
}

@misc{ma2025unitokunifiedtokenizervisual,
      title={UniTok: A Unified Tokenizer for Visual Generation and Understanding}, 
      author={Chuofan Ma and Yi Jiang and Junfeng Wu and Jihan Yang and Xin Yu and Zehuan Yuan and Bingyue Peng and Xiaojuan Qi},
      year={2025},
      eprint={2502.20321},
      archivePrefix={arXiv},
      primaryClass={cs.CV},
      url={https://arxiv.org/abs/2502.20321}, 
}

@article{xiao2023streamingllm,
        title={Efficient Streaming Language Models with Attention Sinks},
        author={Xiao, Guangxuan and Tian, Yuandong and Chen, Beidi and Han, Song and Lewis, Mike},
        journal={arXiv},
        year={2023}
        }

@article{gu2024attention,
      title={When Attention Sink Emerges in Language Models: An Empirical View},
      author={Gu, Xiangming and Pang, Tianyu and Du, Chao and Liu, Qian and Zhang, Fengzhuo and Du, Cunxiao and Wang, Ye and Lin, Min},
      journal={arXiv preprint arXiv:2410.10781},
      year={2024}
}

@article{shang2024LLaVA-PruMerge,
          title={LLaVA-PruMerge: Adaptive Token Reduction for Efficient Large Multimodal Models},
          author={Shang, Yuzhang and Cai, Mu and Xu, Bingxin and Lee, Yong Jae and Yan, Yan},
          journal={arXiv preprint arXiv:2403.15388},
          year={2024}
        }

@misc{llavamini,
      title={LLaVA-Mini: Efficient Image and Video Large Multimodal Models with One Vision Token}, 
      author={Shaolei Zhang and Qingkai Fang and Zhe Yang and Yang Feng},
      year={2025},
      eprint={2501.03895},
      archivePrefix={arXiv},
      primaryClass={cs.CV},
      url={https://arxiv.org/abs/2501.03895}, 
}

@misc{chen2024imageworth12tokens,
      title={An Image is Worth 1/2 Tokens After Layer 2: Plug-and-Play Inference Acceleration for Large Vision-Language Models}, 
      author={Liang Chen and Haozhe Zhao and Tianyu Liu and Shuai Bai and Junyang Lin and Chang Zhou and Baobao Chang},
      year={2024},
      eprint={2403.06764},
      archivePrefix={arXiv},
      primaryClass={cs.CV},
      url={https://arxiv.org/abs/2403.06764}, 
}

@misc{li2024tokenpackerefficientvisualprojector,
      title={TokenPacker: Efficient Visual Projector for Multimodal LLM}, 
      author={Wentong Li and Yuqian Yuan and Jian Liu and Dongqi Tang and Song Wang and Jie Qin and Jianke Zhu and Lei Zhang},
      year={2024},
      eprint={2407.02392},
      archivePrefix={arXiv},
      primaryClass={cs.CV},
      url={https://arxiv.org/abs/2407.02392}, 
}

@misc{hu2024matryoshkaquerytransformerlarge,
      title={Matryoshka Query Transformer for Large Vision-Language Models}, 
      author={Wenbo Hu and Zi-Yi Dou and Liunian Harold Li and Amita Kamath and Nanyun Peng and Kai-Wei Chang},
      year={2024},
      eprint={2405.19315},
      archivePrefix={arXiv},
      primaryClass={cs.CV},
      url={https://arxiv.org/abs/2405.19315}, 
}

@misc{zhang2025avladaptiveattentionlarge,
      title={A-VL: Adaptive Attention for Large Vision-Language Models}, 
      author={Junyang Zhang and Mu Yuan and Ruiguang Zhong and Puhan Luo and Huiyou Zhan and Ningkang Zhang and Chengchen Hu and Xiangyang Li},
      year={2025},
      eprint={2409.14846},
      archivePrefix={arXiv},
      primaryClass={cs.AI},
      url={https://arxiv.org/abs/2409.14846}, 
}

@misc{zhang2025himixreducingcomputationalcomplexity,
              title={HiMix: Reducing Computational Complexity in Large Vision-Language Models},
              author={Xuange Zhang and Dengjie Li and Bo Liu and Zenghao Bao and Yao Zhou and Baisong Yang and Zhongying Liu and Yujie Zhong and Zheng Zhao and Tongtong Yuan},
              year={2025},
              eprint={2501.10318},
              archivePrefix={arXiv},
              primaryClass={cs.CV},
}

@misc{luo2023cheapquickefficientvisionlanguage,
      title={Cheap and Quick: Efficient Vision-Language Instruction Tuning for Large Language Models}, 
      author={Gen Luo and Yiyi Zhou and Tianhe Ren and Shengxin Chen and Xiaoshuai Sun and Rongrong Ji},
      year={2023},
      eprint={2305.15023},
      archivePrefix={arXiv},
      primaryClass={cs.CV},
      url={https://arxiv.org/abs/2305.15023}, 
}

@inproceedings{rao2021dynamicvit,
  title={DynamicViT: Efficient Vision Transformers with Dynamic Token Sparsification},
  author={Rao, Yongming and Zhao, Wenliang and Liu, Benlin and Lu, Jiwen and Zhou, Jie and Hsieh, Cho-Jui},
  booktitle = {Advances in Neural Information Processing Systems (NeurIPS)},
  year = {2021}
}

@misc{bolya2023tokenmergingvitfaster,
      title={Token Merging: Your ViT But Faster}, 
      author={Daniel Bolya and Cheng-Yang Fu and Xiaoliang Dai and Peizhao Zhang and Christoph Feichtenhofer and Judy Hoffman},
      year={2023},
      eprint={2210.09461},
      archivePrefix={arXiv},
      primaryClass={cs.CV},
      url={https://arxiv.org/abs/2210.09461}, 
}

@inproceedings{chen2024sharegpt4v,
  title={Sharegpt4v: Improving large multi-modal models with better captions},
  author={Chen, Lin and Li, Jinsong and Dong, Xiaoyi and Zhang, Pan and He, Conghui and Wang, Jiaqi and Zhao, Feng and Lin, Dahua},
  booktitle={European Conference on Computer Vision},
  pages={370--387},
  year={2024},
  organization={Springer}
}

@article{dubey2024llama3,
  title={The llama 3 herd of models},
  author={Dubey, Abhimanyu and Jauhri, Abhinav and Pandey, Abhinav and Kadian, Abhishek and Al-Dahle, Ahmad and Letman, Aiesha and Mathur, Akhil and Schelten, Alan and Yang, Amy and Fan, Angela and others},
  journal={arXiv e-prints},
  pages={arXiv--2407},
  year={2024}
}

@misc{hudson2019gqanewdatasetrealworld,
      title={GQA: A New Dataset for Real-World Visual Reasoning and Compositional Question Answering}, 
      author={Drew A. Hudson and Christopher D. Manning},
      year={2019},
      eprint={1902.09506},
      archivePrefix={arXiv},
      primaryClass={cs.CL},
      url={https://arxiv.org/abs/1902.09506}, 
}

@inproceedings{
    fu2025mme,
    title={{MME}: A Comprehensive Evaluation Benchmark for Multimodal Large Language Models},
    author={Chaoyou Fu and Peixian Chen and Yunhang Shen and Yulei Qin and Mengdan Zhang and Xu Lin and Jinrui Yang and Xiawu Zheng and Ke Li and Xing Sun and Yunsheng Wu and Rongrong Ji and Caifeng Shan and Ran He},
    booktitle={The Thirty-ninth Annual Conference on Neural Information Processing Systems Datasets and Benchmarks Track},
    year={2025},
    url={https://openreview.net/forum?id=DgH9YCsqWm}
}

@misc{li2023evaluatingobjecthallucinationlarge,
      title={Evaluating Object Hallucination in Large Vision-Language Models}, 
      author={Yifan Li and Yifan Du and Kun Zhou and Jinpeng Wang and Wayne Xin Zhao and Ji-Rong Wen},
      year={2023},
      eprint={2305.10355},
      archivePrefix={arXiv},
      primaryClass={cs.CV},
      url={https://arxiv.org/abs/2305.10355}, 
}

@misc{li2023seedbenchbenchmarkingmultimodalllms,
      title={SEED-Bench: Benchmarking Multimodal LLMs with Generative Comprehension}, 
      author={Bohao Li and Rui Wang and Guangzhi Wang and Yuying Ge and Yixiao Ge and Ying Shan},
      year={2023},
      eprint={2307.16125},
      archivePrefix={arXiv},
      primaryClass={cs.CL},
      url={https://arxiv.org/abs/2307.16125}, 
}

@misc{li2024playgroundv25insightsenhancing,
      title={Playground v2.5: Three Insights towards Enhancing Aesthetic Quality in Text-to-Image Generation}, 
      author={Daiqing Li and Aleks Kamko and Ehsan Akhgari and Ali Sabet and Linmiao Xu and Suhail Doshi},
      year={2024},
      eprint={2402.17245},
      archivePrefix={arXiv},
      primaryClass={cs.CV},
      url={https://arxiv.org/abs/2402.17245}, 
}

@article{sun2023journeydb,
  title={Journeydb: A benchmark for generative image understanding},
  author={Sun, Keqiang and Pan, Junting and Ge, Yuying and Li, Hao and Duan, Haodong and Wu, Xiaoshi and Zhang, Renrui and Zhou, Aojun and Qin, Zipeng and Wang, Yi and others},
  journal={Advances in neural information processing systems},
  volume={36},
  pages={49659--49678},
  year={2023}
}

@inproceedings{
    liu2025world,
    title={World Model on Million-Length Video And Language With Blockwise RingAttention},
    author={Hao Liu and Wilson Yan and Matei Zaharia and Pieter Abbeel},
    booktitle={The Thirteenth International Conference on Learning Representations},
    year={2025},
    url={https://openreview.net/forum?id=HN8V0flwJF}
}

@misc{yu2023scalingautoregressivemultimodalmodels,
      title={Scaling Autoregressive Multi-Modal Models: Pretraining and Instruction Tuning}, 
      author={Lili Yu and Bowen Shi and Ramakanth Pasunuru and Benjamin Muller and Olga Golovneva and Tianlu Wang and Arun Babu and Binh Tang and Brian Karrer and Shelly Sheynin and Candace Ross and Adam Polyak and Russell Howes and Vasu Sharma and Puxin Xu and Hovhannes Tamoyan and Oron Ashual and Uriel Singer and Shang-Wen Li and Susan Zhang and Richard James and Gargi Ghosh and Yaniv Taigman and Maryam Fazel-Zarandi and Asli Celikyilmaz and Luke Zettlemoyer and Armen Aghajanyan},
      year={2023},
      eprint={2309.02591},
      archivePrefix={arXiv},
      primaryClass={cs.LG},
      url={https://arxiv.org/abs/2309.02591}, 
}

@misc{zhan2025anygptunifiedmultimodalllm,
      title={AnyGPT: Unified Multimodal LLM with Discrete Sequence Modeling}, 
      author={Jun Zhan and Junqi Dai and Jiasheng Ye and Yunhua Zhou and Dong Zhang and Zhigeng Liu and Xin Zhang and Ruibin Yuan and Ge Zhang and Linyang Li and Hang Yan and Jie Fu and Tao Gui and Tianxiang Sun and Yu-Gang Jiang and Xipeng Qiu},
      year={2025},
      eprint={2402.12226},
      archivePrefix={arXiv},
      primaryClass={cs.CL},
      url={https://arxiv.org/abs/2402.12226}, 
}

@misc{ge2023makingllamadrawseed,
      title={Making LLaMA SEE and Draw with SEED Tokenizer}, 
      author={Yuying Ge and Sijie Zhao and Ziyun Zeng and Yixiao Ge and Chen Li and Xintao Wang and Ying Shan},
      year={2023},
      eprint={2310.01218},
      archivePrefix={arXiv},
      primaryClass={cs.CV},
      url={https://arxiv.org/abs/2310.01218}, 
}

@misc{jin2024unifiedlanguagevisionpretrainingllm,
      title={Unified Language-Vision Pretraining in LLM with Dynamic Discrete Visual Tokenization}, 
      author={Yang Jin and Kun Xu and Kun Xu and Liwei Chen and Chao Liao and Jianchao Tan and Quzhe Huang and Bin Chen and Chenyi Lei and An Liu and Chengru Song and Xiaoqiang Lei and Di Zhang and Wenwu Ou and Kun Gai and Yadong Mu},
      year={2024},
      eprint={2309.04669},
      archivePrefix={arXiv},
      primaryClass={cs.CV},
      url={https://arxiv.org/abs/2309.04669}, 
}

@misc{lin2025boostingmultimodallargelanguage,
      title={Boosting Multimodal Large Language Models with Visual Tokens Withdrawal for Rapid Inference}, 
      author={Zhihang Lin and Mingbao Lin and Luxi Lin and Rongrong Ji},
      year={2025},
      eprint={2405.05803},
      archivePrefix={arXiv},
      primaryClass={cs.CV},
      url={https://arxiv.org/abs/2405.05803}, 
}

@misc{li2023llamavidimageworth2,
      title={LLaMA-VID: An Image is Worth 2 Tokens in Large Language Models}, 
      author={Yanwei Li and Chengyao Wang and Jiaya Jia},
      year={2023},
      eprint={2311.17043},
      archivePrefix={arXiv},
      primaryClass={cs.CV},
      url={https://arxiv.org/abs/2311.17043}, 
}

@inproceedings{zhuang2025argus,
  title={Argus: A Compact and Versatile Foundation Model for Vision},
  author={Zhuang, Weiming and Chen, Chen and Li, Zhizhong and Sajadmanesh, Sina and Li, Jingtao and Huang, Jiabo and Sehwag, Vikash and Sharma, Vivek and Shinozaki, Hirotaka and Garcia, Felan Carlo and others},
  booktitle={Proceedings of the Computer Vision and Pattern Recognition Conference},
  pages={4418--4429},
  year={2025}
}
}


\end{document}